\documentclass[review]{elsarticle}

\usepackage{graphicx}
\usepackage{booktabs}
\usepackage{amsmath, amssymb}
\usepackage{url}
\usepackage{array}
\usepackage{tabularx}
\usepackage{siunitx}
\usepackage{float}

\usepackage[caption=false,font=footnotesize]{subfig}

\newcolumntype{L}{>{\raggedright\arraybackslash}X}
\newcolumntype{C}{>{\centering\arraybackslash}X}
\newcolumntype{Y}{S[table-format=1.2]}

\journal{Computerized Medical Imaging and Graphics}

\begin{document}

\begin{frontmatter}

\title{Scaling Down to Scale Up: Towards Operationally-Efficient and Deployable Clinical Models via Cross-Modal Low-Rank Adaptation for Medical Vision-Language Models}

\author[inst1]{Thuraya Alzubaidi}
\author[inst2,inst3]{Farhad R. Nezami}
\author[inst1,inst4]{Muzammil Behzad\corref{cor1}}

\cortext[cor1]{Corresponding author.\ead{muzammil.behzad@kfupm.edu.sa}}

\address[inst1]{King Fahd University of Petroleum and Minerals, Saudi Arabia}
\address[inst2]{Institute for Medical Engineering and Science, Massachusetts Institute of Technology, US}
\address[inst3]{Harvard Medical School, Harvard University, US}
\address[inst4]{SDAIA-KFUPM Joint Research Center for Artificial Intelligence, Saudi Arabia
\\
Emails: \url{g202501130@kfupm.edu.sa}, \url{farhadr@mit.edu}, \url{muzammil.behzad@kfupm.edu.sa}
}

\begin{abstract}
Foundation models trained via vision-language pretraining have demonstrated strong zero-shot capabilities across diverse image domains, yet their application to volumetric medical imaging remains limited. We introduce MedCT-VLM: Medical CT Vision-Language Model, a parameter-efficient vision-language framework designed to adapt large-scale CT foundation models for downstream clinical tasks. MedCT-VLM uses a parameter-efficient approach to adapt CT-CLIP, a contrastive vision-language model trained on 25,692 chest CT volumes, for multi-label pathology classification using Low-Rank Adaptation (LoRA). Rather than fine-tuning the model's 440 M parameters directly, we insert low-rank decomposition matrices into attention layers of both vision and text encoders, training only 1.67M parameters (0.38\% of total). We evaluate on zero-shot classification across 18 thoracic pathologies, where the model must align CT embeddings with unseen text prompts at inference without task-specific training. LoRA fine-tuning improves mean AUROC from 61.3\% to 68.9\% (+7.6 pp), accuracy from 67.2\% to 73.6\% (+6.4 pp), and macro-F1 from 32.1\% to 36.9\% (+4.8 pp). These results demonstrate that parameter-efficient methods can effectively transfer large-scale pretraining to downstream medical imaging tasks, particularly for zero-shot scenarios where labeled data is scarce.
\end{abstract}


\end{frontmatter}

\section{Introduction}
Medical imaging analysis increasingly relies on foundation models pretrained on large-scale image-text datasets. Vision-language pretraining learns cross-modal representations by maximizing similarity between matched image-text pairs while minimizing similarity between mismatched pairs. This approach has proven effective for chest X-ray analysis, where models like BioViL, MedCLIP, and CXR-CLIP leverage radiology reports paired with 2D images. The extension to volumetric imaging presents distinct challenges: CT scans contain 200–300 slices per volume, require specialized preprocessing to handle Hounsfield Units, and exhibit greater anatomical complexity than single-view radiographs.

CT-RATE recently addressed these challenges by releasing 25,692 non-contrast chest CT volumes paired with structured radiology reports. The accompanying CT-CLIP model processes 3D volumes through a factorized spatiotemporal transformer and aligns them with radiology text via contrastive learning in a shared 512-dimensional space. While CT-CLIP achieves strong performance on multi-abnormality detection, adapting it to specific clinical tasks traditionally requires full fine-tuning of hundreds of millions of parameters. This is computationally expensive, storage intensive, and risks catastrophic forgetting of pretrained knowledge.

Parameter-efficient fine-tuning methods offer a practical alternative. Low-Rank Adaptation (LoRA) decomposes weight updates into low-rank factors, achieving comparable or better performance than full fine-tuning while training orders of magnitude fewer parameters. CLIP-Adapter and Tip-Adapter demonstrated similar efficiency gains for 2D image classification. However, these methods remain underexplored for volumetric medical imaging, particularly in zero-shot scenarios where the model must align visual features with arbitrary text descriptions without seeing labeled examples during training.

We investigate whether LoRA can improve zero-shot classification in CT-CLIP while maintaining computational efficiency. We inject low-rank adapters into attention layers of both the vision and text encoders, training only 1.67M parameters compared to 440 M for full fine-tuning. Evaluation on 18 thoracic pathologies shows substantial improvements: mean AUROC increases from 61.3\% to 68.9\%, accuracy from 67.2\% to 73.6\%, and macro-F1 from 32.1\% to 36.9\%. The approach reduces checkpoint size by 74× while keeping pretrained weights frozen, enabling efficient multi-task deployment through adapter swapping.

\section{Related Work}

\subsection{Medical Vision-Language Models}
Medical vision-language pretraining learns cross-modal representations by training on paired images and text. This approach has proven effective for medical imaging tasks. Early work in this domain established a great foundation for building research upon. MedCLIP \cite{wang2022medclip} introduced decoupling image and text encoders and replacing standard InfoNCE loss with a semantic matching objective, directly addressing the false negative problem inherent in medical data where different patients frequently exhibit identical findings. The model achieved superior performance with just 20,000 pretraining samples, a tenfold reduction compared to competing approaches that required 200,000 samples. BioViL \cite{boecking2022biovil} demonstrated the value of domain specific text specialization through CXR-BERT, a text encoder focused on radiology tasks that improved cross-modal alignment on chest X-ray tasks including phrase grounding and disease classification. Earlier work like ConVIRT \cite{zhang2020contrastive} and GLoRIA \cite{huang2021gloria} demonstrated the value of contrastive alignment and local and global correspondence, a principle that influenced most medical vision-language approaches that followed.

BioViL-T \cite{bannur2023biovilt} marked a shift toward temporal modeling in medical VLMs. By incorporating prior examinations through a CNN-Transformer hybrid architecture, it achieved state of the art performance on both progression classification and report generation. Temporal modeling addresses how radiologists actually work: comparing current studies against historical baselines is standard practice, and revealed which semantic categories most depend on temporal context from the model’s token level sensitivity analysis, providing interpretable insights into how prior information influences diagnostic reasoning. Recent large scale models such as BioMedCLIP \cite{zhang2023biomedclip} and CheXzero \cite{taylor2022chexzero} further demonstrated gains from scaling dataset size and leveraging report supervision.

The field evolved in response to two constraints: managing linguistic diversity in medical language and working with limited labeled data. Med-UniC \cite{wan2023medunic} tackled cross lingual VLP by introducing Cross lingual Text Alignment Regularization (CTR) to disentangle language related factors while preserving semantic alignment across English and Spanish reports, showing that multimodal representations work across languages. CXR-CLIP \cite{you2023cxrclip} tackled supervision scarcity by converting image labels into pseudo image-text pairs using prompt templates, bridging label-based and report-based training. Study-level contrastive losses (ICL and TCL) allowed simultaneous learning across multiple images and report sections, outperforming comparable methods.

The extension to 3D volumetric imaging marks the current frontier of medical VLP. CT-RATE \cite{hamamci2024ctrate} established the foundation for vision-language models designed specifically for CT imaging, modeling through 25,692 non-contrast chest CT volumes paired with radiology reports. CT-CLIP, the contrastive model trained on this dataset, functions as a task-agnostic vision encoder that outperforms fully supervised baselines on multi-abnormality detection in both internal and external validation. The subsequent development of CT-CHAT (which combines CT-CLIP’s encoder with a pretrained language model fine-tuned on 2.7 million question-answer pairs) demonstrates the viability of conversational interfaces for 3D medical imaging. This progression from 2D chest X-rays to volumetric CT establishes both the architectural patterns and empirical precedent for parameter-efficient adaptation methods such as VL-Adapter \cite{sung2022vladapter}, CLIP-Adapter \cite{gao2024clipadapter}, Tip-Adapter \cite{zhang2022tipadapter}, CLIP-LoRA \cite{zanella2024cliplora}, Proto-Adapter \cite{kato2024protoadapter}, and MMA \cite{yang2024mma} in medical VLP systems.

\begin{table*}[!htbp]
\centering
\caption{Summary of key medical vision--language pretraining models}
\label{tab:med_vlp_section1}
\scriptsize
\setlength{\tabcolsep}{3pt}
\begin{tabular}{p{0.12\textwidth}|p{0.20\textwidth}|p{0.15\textwidth}|p{0.25\textwidth}|p{0.25\textwidth}}
\toprule
\textbf{Model} & \textbf{Dataset(s)} & \textbf{Tasks} & \textbf{Methodology (Key Point)} & \textbf{Results (Key Finding)} \\
\midrule

\textbf{MedCLIP (2022) \cite{wang2022medclip}} &
Paired medical image--text datasets (MIMIC-CXR used for evaluation). &
Zero-shot classification, supervised classification, retrieval. &
Decouples image/text encoders; replaces InfoNCE with semantic matching loss to mitigate false negatives in medical imaging. &
Achieves stronger performance than prior methods while using only $\sim$20k pretraining pairs (vs. 200k in baselines). \\

\midrule

\textbf{BioViL (2022) \cite{boecking2022biovil}} &
MIMIC-CXR, MS-CXR (phrase-grounding annotations). &
Phrase grounding, report–image alignment, classification. &
Uses a radiology-specialized text encoder (CXR-BERT) and multi-level contrastive alignment across tokens, regions, and sentences. &
Improves cross-modal alignment; sets strong benchmarks on grounding and CXR classification tasks. \\

\midrule

\textbf{BioViL-T (2023) \cite{bannur2023biovilt}} &
MS-CXR-T (temporal extension of MS-CXR), MIMIC-CXR. &
Progression classification, phrase grounding, report generation. &
Incorporates temporal priors using a CNN--Transformer multi-image encoder with explicit modeling of prior exams. &
Achieves SOTA on progression classification and grounding; demonstrates that temporal cues significantly enhance diagnostic reasoning. \\

\midrule

\textbf{Med-UniC (2023) \cite{wan2023medunic}} &
MIMIC-CXR (English), PadChest (Spanish). &
Cross-lingual retrieval, classification, grounding. &
Introduces Cross-lingual Text Alignment Regularization (CTR) to align English and Spanish reports while suppressing language-specific biases. &
Outperforms monolingual models across 5 tasks and 10 datasets; demonstrates robust cross-lingual medical VLP. \\

\midrule

\textbf{CXR-CLIP (2023) \cite{you2023cxrclip}} &
Large-scale chest X-ray dataset with labels + reports (from hospital PACS + public sources). &
CXR classification and retrieval. &
Converts image–label pairs into pseudo image–text pairs using prompt templates; uses study-level contrastive losses (ICL/TCL). &
Outperforms prior VLP models trained under identical conditions; study-level training improves classification across datasets. \\

\midrule

\textbf{CT-CLIP (2024) \cite{hamamci2024ctrate}} &
CT-RATE (25{,}692 non-contrast chest CT volumes + reports). &
Zero-shot multi-abnormality detection, retrieval. &
3D vision–language contrastive pretraining using volumetric CT inputs and paired radiology reports. &
Outperforms fully supervised CT-Net on internal and external validation sets for multi-abnormality detection and retrieval. \\

\bottomrule
\end{tabular}
\end{table*}

\subsection{Adapter-Based and PEFT Approaches}
Large-scale VLMs are computationally expensive to fine-tune fully. Parameter-efficient fine-tuning (PEFT) offers an alternative. VL-Adapter \cite{sung2022vladapter} showed that adapter modules can effectively adapt vision-language transformers to new tasks. By inserting small adapter layers into VL-T5 and VL-BART, the approach matched full fine-tuning performance on diverse image-text and video-text tasks. Weight sharing kept parameter usage to 4--6\% of the total. Adapters proved viable for scaling VLMs to multiple tasks simultaneously. In medical imaging, where resources and labeled data are constrained, this capability is particularly relevant.

CLIP's zero-shot strength makes it appealing for adaptation. CLIP-Adapter \cite{gao2024clipadapter} took a straightforward approach, bottleneck layers on visual and text streams, with residual connections preserving original CLIP features while the backbone stayed frozen. Few-shot experiments showed this beat prompt-learning methods like CoOp, validating feature adaptation over input-space tuning.

Building on feature adaptation, Tip-Adapter \cite{zhang2022tipadapter} introduced training-free adapters by caching key-value pairs from few-shot samples where CLIP visual features were keys and labels were values. Inference relied on feature retrieval, with optional brief fine-tuning of adapter weights to improve performance while maintaining minimal overhead.

PEFT strategies beyond bottleneck adapters have emerged. CLIP-LoRA \cite{zanella2024cliplora} examined low-rank adaptation across CLIP's vision and text encoders, exploring which matrices to adapt and decomposition ranks. Results showed LoRA could match or exceed bottleneck adapters with stable cross-dataset performance—a distinct efficiency–capacity tradeoff. Proto-Adapter \cite{kato2024protoadapter} simplified the training-free approach by using prototypes from few-shot data instead of learned weights, matching Tip-Adapter's few-shot performance with less complexity. MMA \cite{yang2024mma} advanced adapter design by modeling cross-modal interactions explicitly rather than treating modalities separately, boosting discrimination and robustness in few-shot scenarios. Collectively, these approaches trace a path from basic bottleneck adapters toward methods that either skip training entirely or capitalize on multimodal structure. This diversity of strategies makes adapting foundation models like CT-CLIP to medical imaging tasks more flexible. 

\subsection{Advances in VLMs Methods for Radiology Report Generation}
Report generation moved beyond templates to neural approaches. CAMANet \cite{wang2024camanet} introduced explicit visual attention guidance using Class Activation Maps. Three modules work together: one generates visual discriminative maps to weight tokens, another enhances discriminative features with CAM assistance, and a third aligns attention distributions across modalities. Testing on IU X-Ray and MIMIC-CXR showed this explicit alignment beats implicit learned representations in report quality.

End-to-end generation faces competition from structured alternatives. Replace and Report \cite{kale2023replace} reframes report generation as template editing: a multilabel classifier predicts image tags, a transformer generates pathological descriptions from tags, and a BERT classifier identifies template spans for replacement. On IU X-Ray, it achieved substantial gains (25\% BLEU-1, 36\% ROUGE-L, 44\% METEOR, 48\% CIDEr), showing structured generation through template manipulation outperforms direct image-to-text approaches.

The integration of large language models has introduced new opportunities for parameter-efficient report generation. R2GenGPT \cite{wang2023r2gengpt} aligned visual features into frozen LLM word embedding spaces through a trainable visual alignment module, achieving state-of-the-art performance while training only approximately 5 million parameters (0.07\% of total model parameters). This delta tuning strategy exemplifies PEFT applied to the language modeling component, complementing vision-side adapter methods. SERPENT-VLM \cite{kapadnis2024serpent} addressed hallucination through self-refinement, introducing a self-supervised loss based on similarity between pooled image representations and contextual representations of generated text. Evaluated on IU X-Ray and ROCO against baselines including LLaVA-Med and BiomedGPT, the method demonstrated both improved performance and robustness to noisy images, highlighting the importance of post-hoc alignment for reliable clinical text generation.

Two recent surveys map the growing field. Sloan et al. \cite{sloan2024automated} reviewed datasets, training methods (contrastive and reinforcement learning), model architectures (CNN-RNN, Transformers, multimodal variants), knowledge integration, and evaluation approaches. Wang et al. \cite{wang2025survey} proposed a five-stage workflow: data acquisition, preparation, feature learning, fusion and interaction, and generation. Both surveys highlight the same gap: while NLP metrics work for benchmarking, clinical validation and robustness remain underdeveloped. This matters especially for 3D volumetric imaging, where research lags far behind 2D chest X-ray applications.

\section{Dataset and Preprocessing}
\subsection{Dataset}

We use CT-RATE. While the full dataset contains more than 50\,k CT volumes, we sample \textbf{1,000} 3D chest CT scans with paired radiology reports. Each scan is stored in NIfTI (.nii.gz) format with variable depth (typically \textbf{200–300} slices, minimum \textbf{20}) and fixed spatial dimensions of \textbf{480×480 px}. The dataset includes multi-label pathology annotations across 18 thoracic conditions: medical material, arterial wall calcification, cardiomegaly, pericardial effusion, coronary artery wall calcification, hiatal hernia, lymphadenopathy, emphysema, atelectasis, lung nodule, lung opacity, pulmonary fibrotic sequela, pleural effusion, mosaic attenuation pattern, peribronchial thickening, consolidation, bronchiectasis, and interlobular septal thickening.

\begin{figure*}[t]
    \centering
    \includegraphics[width=\textwidth]{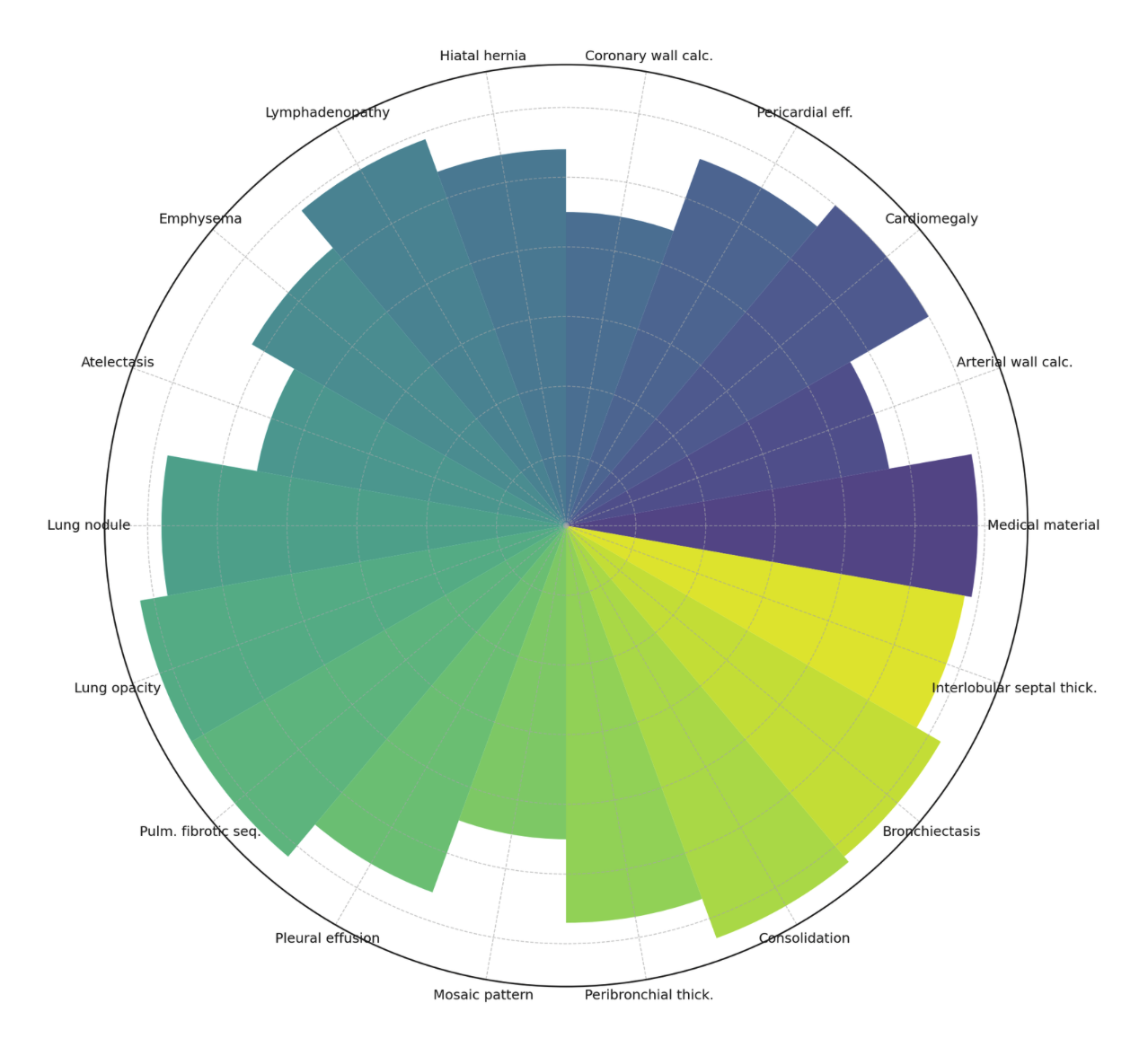}
    \caption{Dataset class distribution across 18 thoracic pathologies.}
    \label{fig:data dist}
\end{figure*}

Each radiology report contains structured sections (Clinical Information, Technique, Findings, Impressions), with the Findings and Impressions sections providing the primary diagnostic content. Reports are tokenized to a maximum of 256 tokens using the BiomedVLP-CXR-BERT tokenizer. In this regard, Figure~\ref{fig:data dist} shows the class distribution across all 18 thoracic pathologies in the sampled subset.

\subsection{Preprocessing Pipeline}

CT volumes are loaded using nibabel and undergo the following preprocessing: (1) application of RescaleSlope and RescaleIntercept from metadata, (2) HU clipping to $[-1000, 1000]$, (3) depth adjustment to 240 slices through resampling or zero-padding, (4) normalization to $[0,1]$, and (5) conversion to tensor format $(1, 240, 480, 480)$. 

\subsection{Data Augmentation}

We apply augmentation transforms in two stages: spatial transforms applied jointly to image and mask (trilinear and nearest-neighbor interpolation respectively), followed by intensity transforms on images only. All spatial transforms preserve anatomical alignment, with flips disabled along the superior-inferior axis to maintain diagnostic orientation. This augmentation pipeline is summarized in Table~\ref{tab:augmentation}.

\begin{table}[!t]
\centering
\caption{CT volume augmentation parameters.}
\label{tab:augmentation}
\scriptsize
\begin{tabular}{p{3cm} p{3.5cm} p{1cm}}
\toprule
\textbf{Augmentation} & \textbf{Configuration} & \textbf{Probability} \\
\midrule
Rotation & $\pm15°$ rotation around random axes & 0.9 \\
Scaling & Zoom in/out by 0.9--1.1× & 0.9 \\
Translation & Shift up to 10~mm in random direction & 0.9 \\
Elastic Deformation & $7 \times 7 \times 7$ grid with max 10~mm displacement & 0.7 \\
Flip & Left-right and anterior-posterior axes & 0.5 \\
Blur & Gaussian smoothing ($\sigma = 0.5$--1.5 voxels) & 0.5 \\
Noise & Gaussian noise ($\sigma = 0.15$ post-normalization) & 0.5 \\
Gamma Correction & Brightness adjustment ($\log(\gamma) \in [-0.5, 0.5]$) & 0.5 \\
Bias Field & Low-frequency intensity variation (coefficient = 0.5) & 0.3 \\
\bottomrule
\end{tabular}
\end{table}

Elastic deformation simulates anatomical variability by applying spatially-varying displacement fields across the volume. We use B-spline interpolation to generate smooth, realistic deformations:
\begin{equation}
\mathbf{x}' = \mathbf{x} + \mathbf{u}(\mathbf{x}), \quad \mathbf{u}(\mathbf{x}) = \sum_{i,j,k} \mathbf{c}_{ijk} B_i(x) B_j(y) B_k(z)
\end{equation}
where $B_\cdot$ are cubic B-spline basis functions and $\mathbf{c}_{ijk}$ control the displacement magnitude. Displacements are reduced at image edges to prevent unrealistic warping.

Gamma correction adjusts image brightness by rescaling voxel values as $I' = I^\gamma$, where $\gamma = e^u$ with $u \sim \mathcal{U}[-0.5, 0.5]$. We keep this range narrow to preserve the clinical validity of Hounsfield Units. The bias field mimics intensity variations seen in CT scanners by multiplying the volume by a smooth, low-order polynomial. This adds realistic scanner artifacts without destroying fine anatomical structures.

\begin{figure*}[t]
    \centering
    \includegraphics[width=\textwidth]{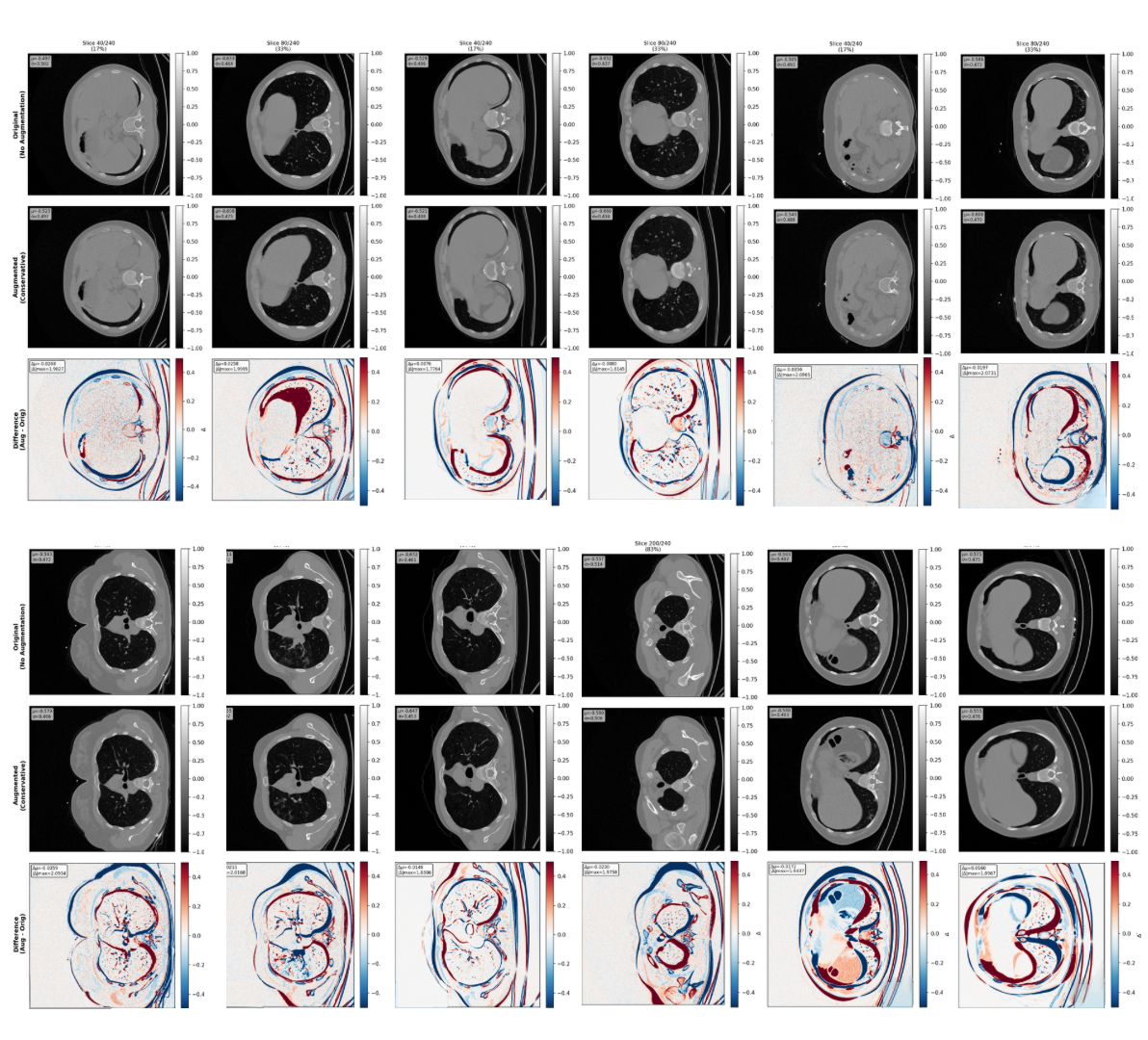}
    \caption{Random samples of CT slices before and after augmentation}
    \label{fig:aug_res}
\end{figure*}

We implemented all augmentations using TorchIO v0.19.6 with fixed random seeds for reproducibility. In Figure~\ref{fig:aug_res}, we illustrate random augmentation results, showing the spatial and intensity variability introduced during preprocessing.

\subsection{Text Augmentation}

For text preprocessing, we concatenate the Findings and Impressions sections from each radiology report, which contain the primary diagnostic information. Reports are tokenized using the BiomedVLP-CXR-BERT tokenizer with padding to 256 tokens maximum. The tokenization process produces input\_ids and attention\_mask tensors for model input.

To increase training diversity without compromising clinical accuracy, we applied minimal text augmentation by replacing select words with clinically equivalent synonyms generated using Claude Sonnet. Replacements preserve diagnostic meaning while introducing lexical variation. We avoided aggressive augmentation techniques such as back-translation or paraphrasing that risk altering the clinical intent of reports.

\section{MedCT-VLM: The Proposed Medical CT Vision-Language Model}
In the following sections, we introduce our proposed MedCT-VLM, which augments this foundational framework with parameter-efficient CrossModal-LoRA modules to further enhance multimodal alignment.

\subsection{Overview}

\begin{figure*}[t]
    \centering
    \includegraphics[width=\textwidth]{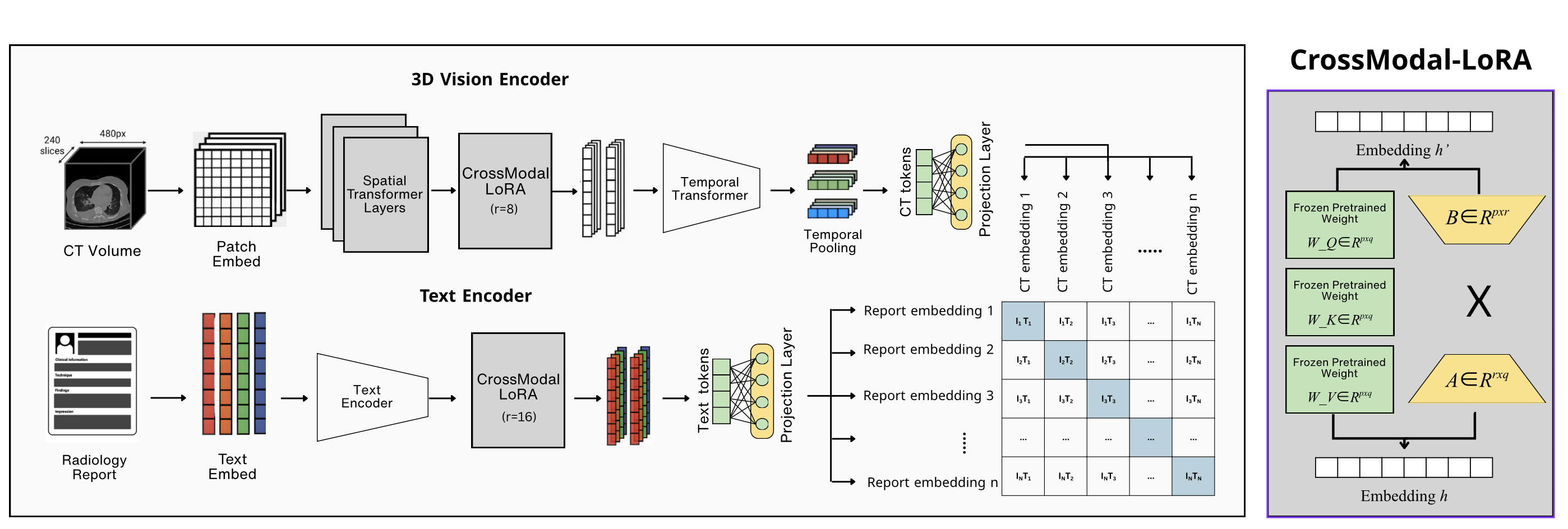}
    \caption{High-level architecture of the proposed model.}
    \label{fig:arch}
\end{figure*}

We utilized a base model with a dual-encoder contrastive learning framework that aligns 3D chest CT volumes with radiology reports through a shared 512-dimensional latent space. The architecture comprises a CTViT (CT Vision Transformer) for processing volumetric scans and a BiomedVLP-CXR-BERT encoder for radiology text. The model learns multimodal representations by increasing the similarity between each CT volume and its corresponding report while decreasing similarity to all unrelated reports. This bidirectional contrastive objective encourages matched CT–text pairs to cluster tightly together in embedding space while pushing mismatched pairs apart, effectively teaching the model how clinical findings in imaging relate to their radiological descriptions. In Figure~\ref{fig:arch}, we provide a high-level illustration of the overall architecture, including the 3D vision encoder, text encoder, and the CrossModal-LoRA modules.

\subsection{Vision Encoder}
CTViT processes 3D CT volumes efficiently using factorized spatiotemporal attention, which decomposes full 3D attention into separate spatial and temporal operations. This reduces computational cost from $O(D^2H^2W^2)$ to $O(D^2 + H^2W^2)$, where $D$, $H$, and $W$ represent depth, height, and width respectively.

Input volumes of size $1 \times D \times 480 \times 480$ are divided into patches: spatial patches and temporal patches of 10 slices. with $24 \times 24$ spatial patches per slice. The encoder then processes these patches through two sequential transformer stacks. First, a 4-layer spatial transformer with 8 attention heads and 32-dimensional head size extracts features within each slice. Second, a 4-layer temporal transformer captures dependencies across slices, building 3D anatomical continuity.

Following the transformers, a vector quantization module with 8,192 codebook entries compresses the learned representations into discrete codes. The resulting 294,912-dimensional embedding is then projected to 512 dimensions for downstream tasks.

\subsection{Text Encoder}

Radiology reports are encoded using BiomedVLP-CXR-BERT, a BERT model pretrained on chest X-ray reports with an extended vocabulary for medical terminology. The encoder processes tokenized text (capped at 256 tokens) through 12 transformer layers. The [CLS] token representation, a standard approach for capturing document-level meaning, is extracted as a 768-dimensional embedding. This embedding is then projected to 512 dimensions to align with visual features in the shared representation space.

\subsection{Cross-Modal Parameter-Efficient Fine-Tuning with LoRA}

Full fine-tuning of the base model's 440 M parameters presents significant challenges: high computational cost, large storage requirements ($\sim$500 MB per checkpoint), and risk of catastrophic forgetting of pretrained knowledge. To enable efficient domain adaptation and task-specific tuning, we employ LoRA (Low-Rank Adaptation), training only 0.38\% of model parameters (1.67M) while keeping the base model frozen.

\subsubsection{LoRA Formulation}

For a pretrained weight matrix $W_0 \in \mathbb{R}^{d \times k}$,
LoRA decomposes weight updates into low-rank factors:
\begin{equation}
W = W_0 + \Delta W = W_0 + BA
\end{equation}
where $B \in \mathbb{R}^{d \times r}$ and $A \in \mathbb{R}^{r \times k}$ are trainable matrices with rank $r \ll \min(d,k)$. Since $r$ is much smaller than the original dimensions, this drastically reduces the number of trainable parameters.

During forward pass, the adapted weights are applied as:
\begin{equation}
h = W_0x + \frac{\alpha}{r}BAx
\end{equation}
The scaling factor $\frac{\alpha}{r}$ controls the strength of the adaptation relative to the pretrained weights. Matrix $A$ is randomly initialized using Kaiming uniform sampling, while $B$ starts at zero. This initialization ensures $\Delta W = \mathbf{0}$ initially, so the model behaves identically to the pretrained version before training begins.

\subsubsection{Adapter Configuration}

LoRA adapters are injected into attention projection layers of both encoders. For the vision encoder, we apply adapters to query, key-value, and output projections in both spatial and temporal transformers. We use rank $r_{\text{vision}} = 16$, scaling factor $\alpha_{\text{vision}} = 32.0$, and dropout $p = 0.05$, totaling 97 adapter layers. For the text encoder (BERT), adapters target the query, key, value, and dense layers across all 12 transformer blocks with rank $r_{\text{text}} = 8$ and scaling factor $\alpha_{\text{text}} = 16.0$.

\begin{table}[!b]
\centering
\caption{LoRA parameter efficiency: trainable parameters vs.\ model size.}
\label{tab:lora_efficiency}
\scriptsize
\begin{tabular}{p{2.8cm} p{1.8cm} p{1.8cm} p{1.8cm}}
\toprule
\textbf{Component} & \textbf{Model Size} & \textbf{LoRA Params} & \textbf{Trainable \%} \\
\midrule
Vision Encoder & 98M & 1.2M & 1.22\% \\
Text Encoder & 110M & 0.47M & 0.43\% \\
Projection Heads & 232M & 0 & 0.00\% \\
\midrule
\textbf{Total} & \textbf{440M} & \textbf{1.67M} & \textbf{0.38\%} \\
\bottomrule
\end{tabular}
\end{table}

This configuration reduces checkpoint sizes enabling storage of multiple task-specific adapters while sharing a single base model. Adapters can be merged into base weights at inference for zero latency overhead or swapped dynamically for multi-task deployment.

\subsection{Classification Head}
For multi-label pathology classification, we extend the vision encoder by adding a linear classification head that maps the 512-dimensional projected embeddings to 18 output dimensions (512 $\rightarrow$ 18 dimensions). To improve generalization and mitigate overfitting, we include a dropout layer with a dropout probability of ($p=0.3$) before the final linear layer. This classification head operates directly on the encoder’s learned representations to produce the multi-label predictions.

\subsection{Fine-Tuning Strategy}

We fine-tune using LoRA adapters, keeping all pretrained weights $W_0$ frozen while training only the low-rank matrices $(A, B)$ and the classification head. The model is optimized for 15 epochs using AdamW with learning rate $\eta = 5 \times 10^{-4}$ and weight decay $\lambda = 0.01$.

Multi-label classification is trained via binary cross-entropy loss:
\begin{equation}
\mathcal{L}_{\text{BCE}} = -\frac{1}{N \cdot C}\sum_{i=1}^{N}\sum_{j=1}^{C} \left[y_{ij} \log(\sigma(z_{ij})) + (1-y_{ij})\log(1-\sigma(z_{ij}))\right]
\end{equation}
where $N$ is batch size, $C=18$ pathology classes, $y_{ij} \in \{0, 1\}$ denotes ground truth labels, and $z_{ij}$ represents predicted logits. We use BCE instead of softmax cross-entropy because it handles co-occurring pathologies in multi-label scenarios.

\subsection{Model Core Tasks}

\textbf{Multi-Label Classification.} The primary task is thoracic pathology detection: classifying 18 disease types from CT volumes. Direct fine-tuning on this task yields the strongest performance improvements, as adapters learn to extract disease-specific visual patterns.

\textbf{Zero-Shot Detection.} Zero-shot pathology detection leverages text-image alignment by computing similarity scores $p_j = \sigma(\text{sim}(v, t_j))$ between CT embeddings $v$ and pathology text embeddings $\{t_j\}_{j=1}^{C}$ (e.g., ``CT scan showing pneumothorax''). Fine-tuning provides modest gains here, as the task benefits from improved visual representations without being directly optimized.

\textbf{Retrieval.} Volume-to-volume and report-to-volume retrieval tasks rank similar scans and enable cross-modal search using cosine similarity in the shared embedding space. Evaluation metrics are Recall@K and Mean Reciprocal Rank (MRR). These tasks depend primarily on anatomical similarity from pretraining rather than classification-specific features.

\section{Experimental Setup and Evaluation}
In this section, we outline the experimental setup used to evaluate our proposed approach and detail the specific tasks considered in our study.

\subsection{Zero-Shot Classification}

Zero-shot pathology classification is the primary focus because it isolates improvements in visual representation quality from task-specific optimization. Unlike supervised classification, zero-shot performance directly measures transfer capability: the model must leverage better embeddings to align with unseen text prompts at inference time.

\subsection{Zero-Shot Inference Protocol}
For each of the 18 pathologies, we create a text prompt using the template ``CT scan showing \{pathology\}'' (e.g., ``CT scan showing pneumothorax''). The text encoder produces embeddings $\mathbf{t}_j \in \mathbb{R}^{768}$, which are projected to the shared 512-dimensional space: $\mathbf{t}_j' = W_{\text{text}} \mathbf{t}_j$. The CT volume is similarly encoded and projected to $\mathbf{v} \in \mathbb{R}^{512}$.

The predictions are computed by measuring similarity between CT and text embeddings:
\begin{equation}
p_j = \sigma\left(\frac{\mathbf{v}^\top \mathbf{t}_j'}{\|\mathbf{v}\| \|\mathbf{t}_j'\|} \cdot \tau\right)
\end{equation}
where $\sigma(\cdot)$ is the sigmoid function and $\tau$ is a temperature parameter controlling prediction confidence. No classification heads or labeled examples are required at inference.

\subsection{Evaluation Metrics}

We report Area Under the Receiver Operating Characteristic curve (AUROC) and mean AUROC across all pathology classes. Additional multi-label metrics include accuracy, micro-F1, macro-F1, weighted-F1, and samples-F1.

AUROC is threshold-independent and well-suited for imbalanced medical data. Accuracy measures overall correct predictions:
\begin{equation}
\text{Accuracy} = \frac{TP + TN}{TP + TN + FP + FN}
\end{equation}
Micro-F1 aggregates true and false positives across all classes:
\begin{equation}
\text{Micro-F1} = \frac{2 \sum TP}{2 \sum TP + \sum FP + \sum FN}
\end{equation}
Macro-F1 computes F1 per class then averages:
\begin{equation}
\text{Macro-F1} = \frac{1}{C}\sum_{j=1}^{C} F1_j
\end{equation}
where $C=18$ pathology classes and $F1_j = \frac{2 \cdot \text{Precision}_j \cdot \text{Recall}_j}{\text{Precision}_j + \text{Recall}_j}$ for each class.
\section{Results}
We now report the results of our experiments, followed by a detailed analysis of the model’s performance.

\subsection{Overall Zero-Shot Classification Performance}

\begin{figure}[!b]
\centering
\includegraphics[width=\linewidth]{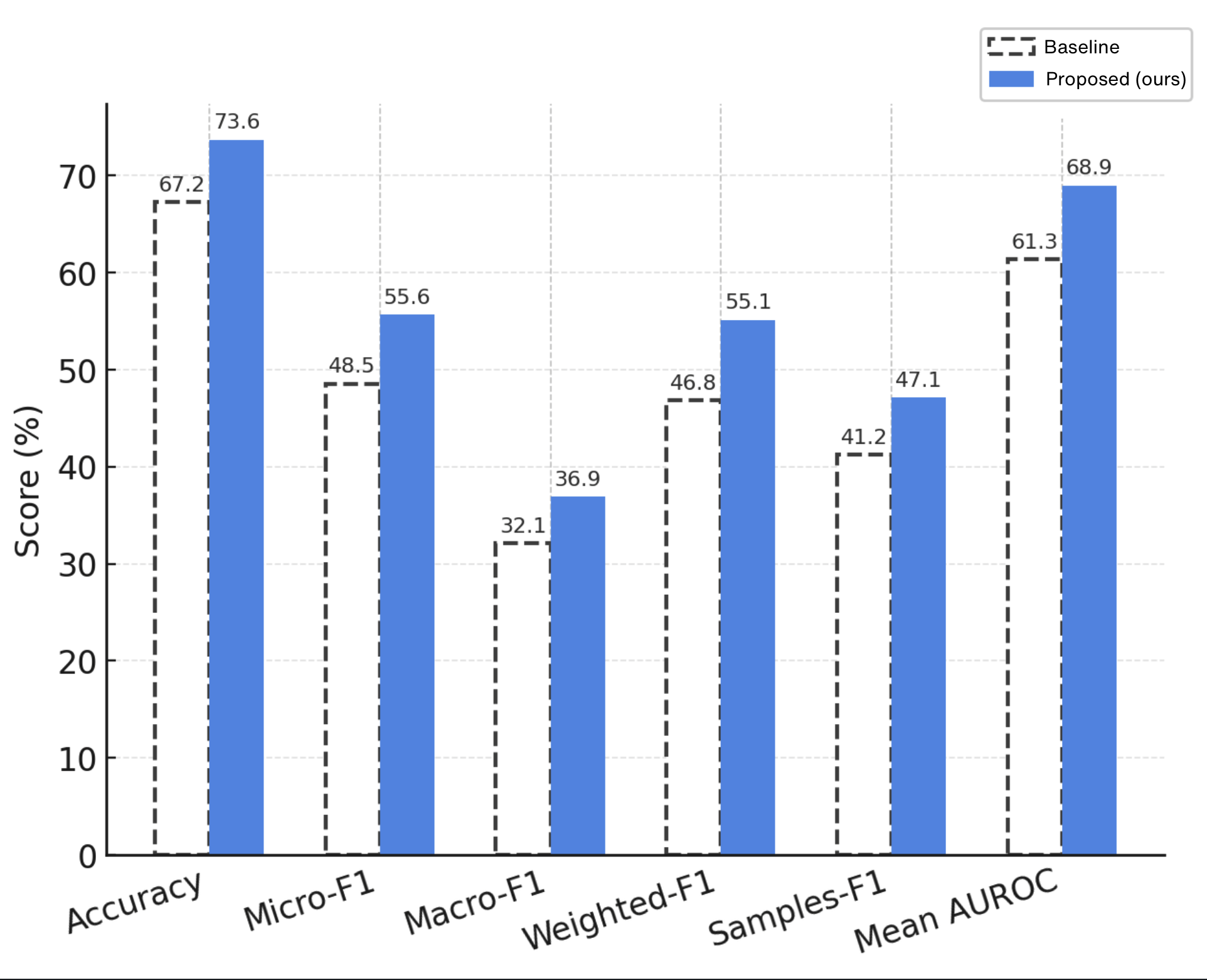}
\caption{Overall zero-shot metrics for Base model vs.\ MedCT-VLM.}
\label{fig:barplot}
\end{figure}

\begin{figure}[!b]
\centering
\includegraphics[width=\linewidth]{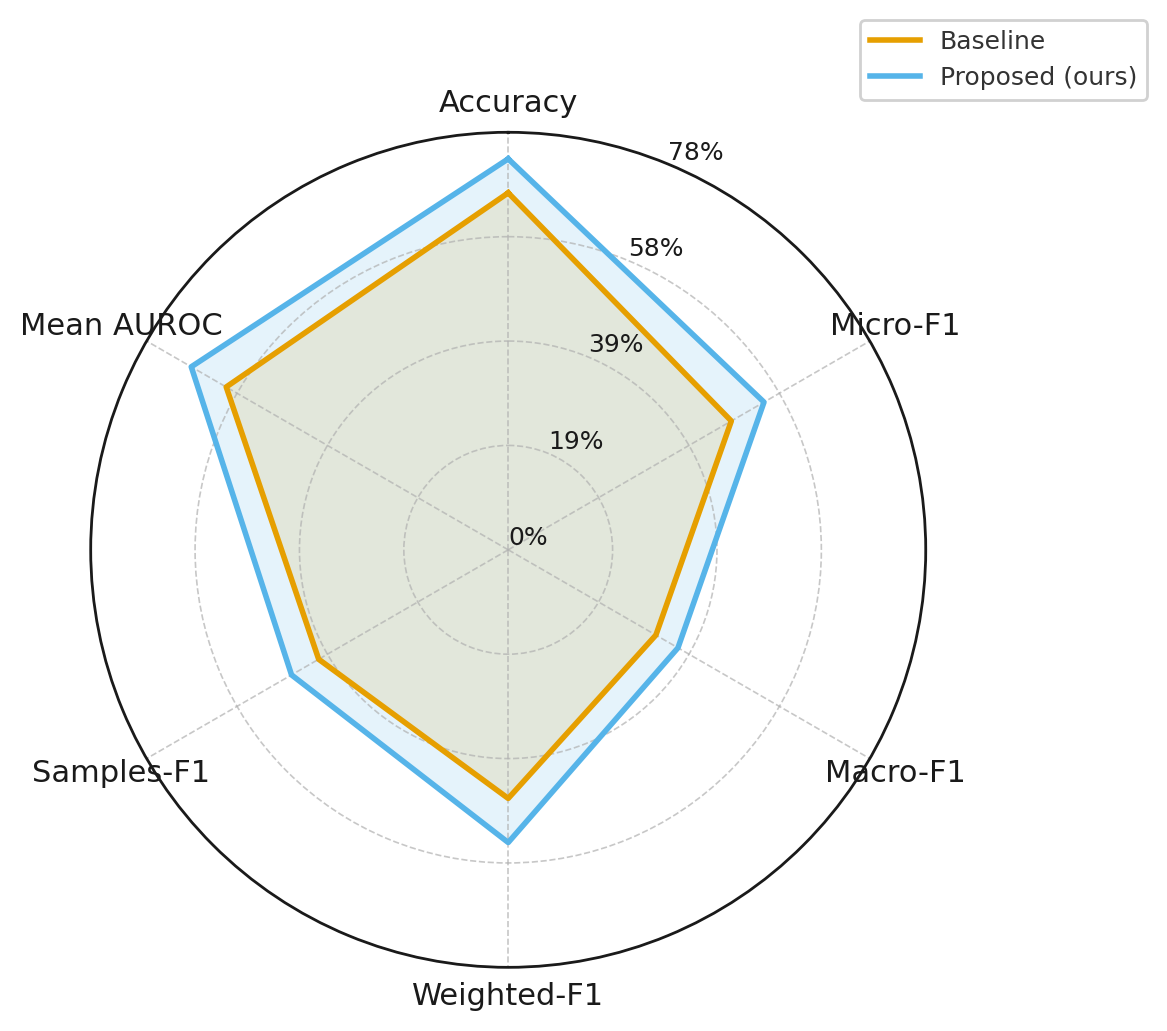}
\caption{Overall metrics (radar): proposed model outperforms baseline across accuracy, F1 variants, samples-F1, and mean AUROC.}
\label{fig:radar}
\end{figure}

\begin{figure}[!t]
\centering
\setlength{\tabcolsep}{2pt}
\begin{tabular}{@{}cc@{}}
  \includegraphics[width=0.49\columnwidth]{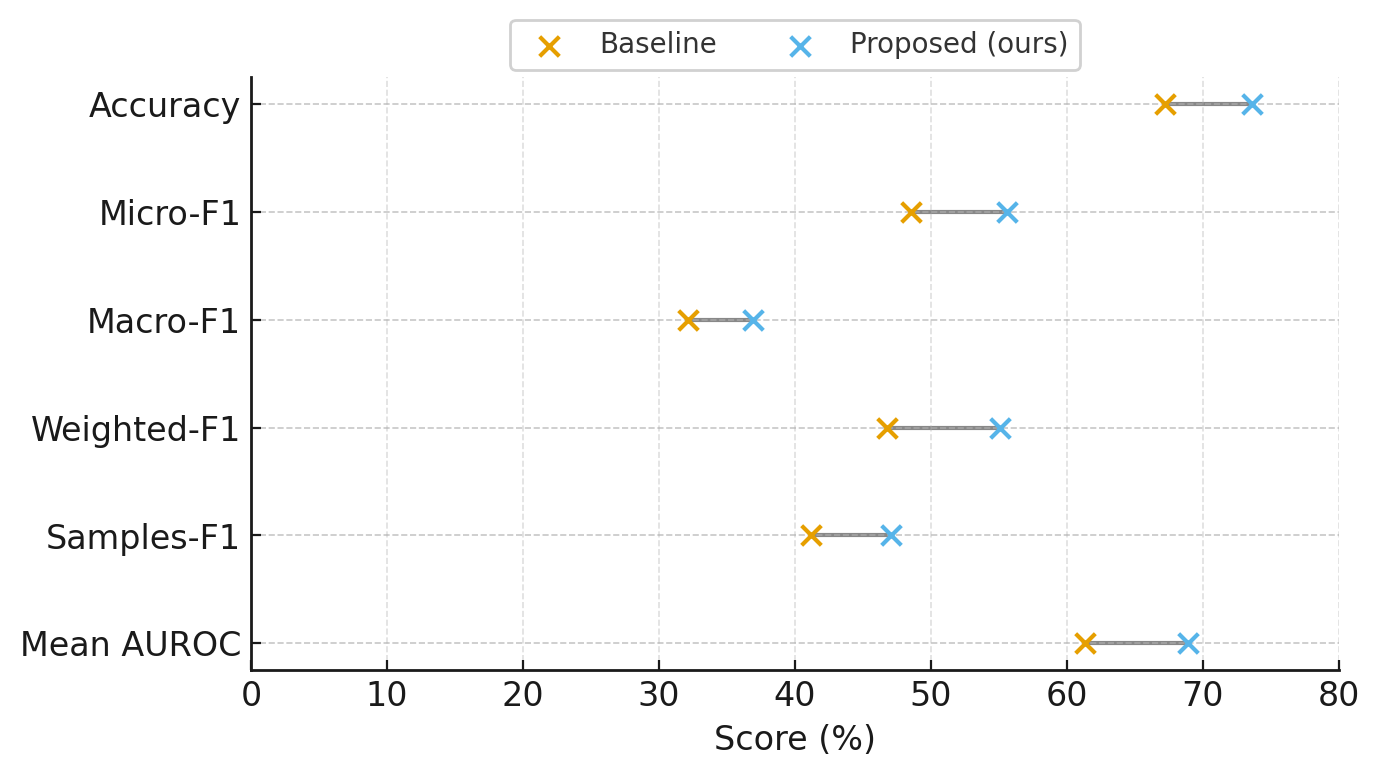} &
  \includegraphics[width=0.49\columnwidth]{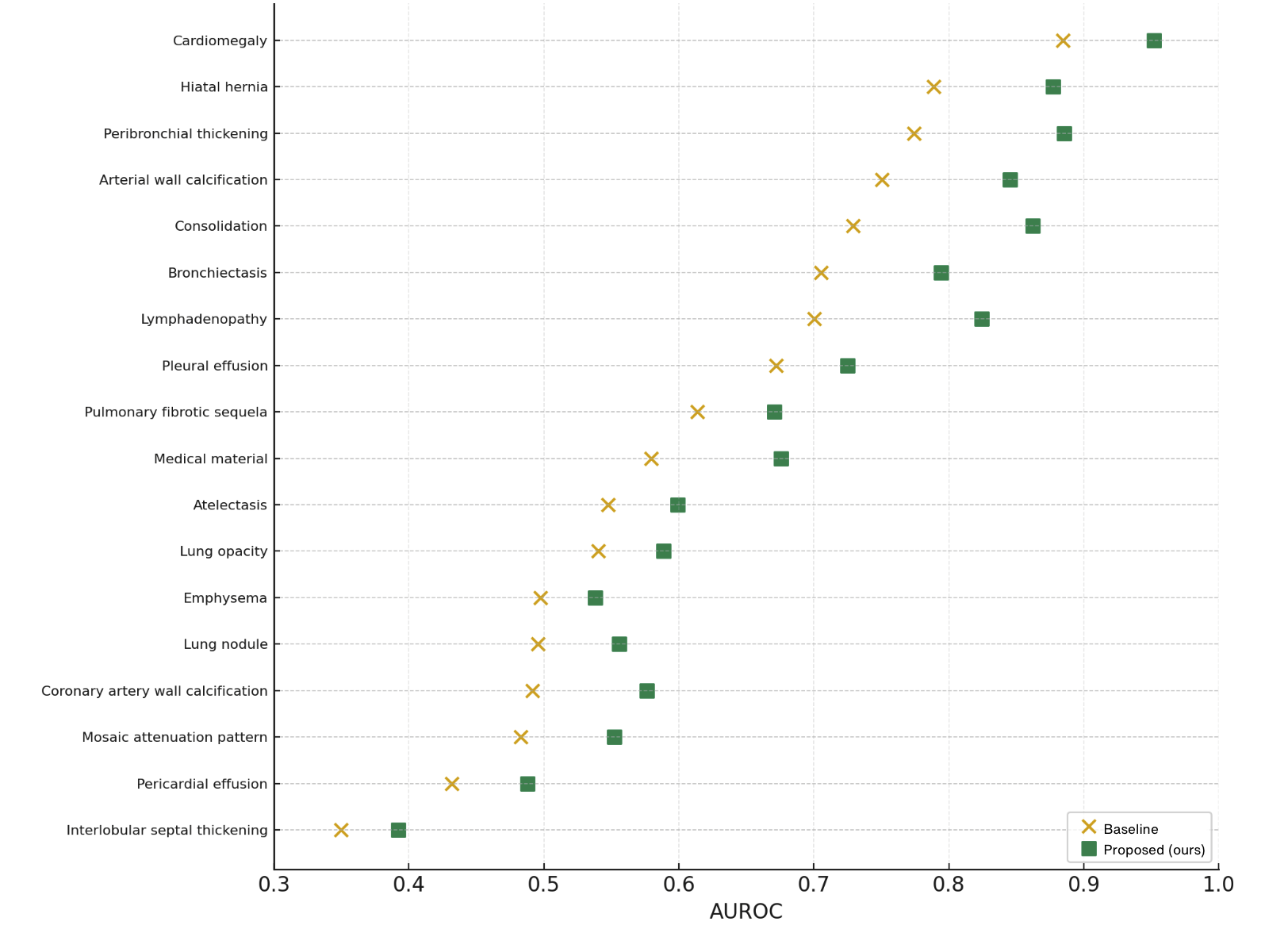} \\
  \small (a) Overall metrics & \small (b) Per-pathology AUROC
\end{tabular}
\caption{Side-by-side comparison of overall and per-pathology results.}
\label{fig:side_by_side}
\end{figure}

In Figures~\ref{fig:barplot}--\ref{fig:side_by_side}, we collectively illustrate the performance improvements achieved by MedCT-VLM across multiple evaluation dimensions. To provide clearer emphasis and avoid conflating their individual contributions, we describe each figure in a dedicated paragraph below.

In Figure~\ref{fig:barplot}, we present the overall zero-shot classification metrics for the base model and MedCT-VLM. The barplot highlights consistent gains across accuracy, F1-based metrics, and mean AUROC, visually reinforcing the numerical improvements reported in Table~\ref{tab:overall_metrics}. This figure provides the clearest high-level comparison and demonstrates that MedCT-VLM delivers uniform benefits regardless of metric choice.

Similarly, Figure~\ref{fig:radar} complements the barplot by offering a holistic view of model behaviour across metrics within a single radial layout. The radar plot makes it evident that MedCT-VLM dominates the baseline across all axes, including accuracy, micro-/macro-F1, weighted-F1, samples-F1, and AUROC. This visualization underscores not only improvement magnitude but also the balanced nature of the gains, showing that no dimension is disproportionately weak or left behind.

Additionally, Figure~\ref{fig:side_by_side} provides a more granular perspective by juxtaposing overall metric improvements with per-pathology AUROC changes. Panel (a) reiterates general performance boosts, while panel (b) highlights pathology-specific benefits. The per-pathology AUROC plot demonstrates that MedCT-VLM improves discriminative capability across nearly all conditions, confirming that the model’s enhancements extend beyond aggregate metrics to clinically relevant, fine-grained decision boundaries.

In Table~\ref{tab:overall_metrics}, we show that our proposed model consistently improves zero-shot classification performance. Accuracy increases from 67.2\% to 73.6\% (+6.4 pp), while macro-F1 improves from 32.1\% to 36.9\% (+4.8 pp). Mean AUROC increases from 61.3\% to 68.9\% (+7.6 pp), demonstrating that fine-tuning enhances visual representation quality and transferability to zero-shot scenarios.

\begin{table}[!b]
\centering
\caption{Zero-shot classification performance: comprehensive metric comparison.}
\label{tab:overall_metrics}
\scriptsize
\begin{tabular}{p{2.5cm} p{1.6cm} p{1.6cm} p{1.4cm} p{1.5cm}}
\toprule
\textbf{Metric} & \textbf{Base model} & \textbf{MedCT-VLM} & \textbf{Gain (pp)} & \textbf{Relative} \\
\midrule
Accuracy & 67.2\% & 73.6\% & +6.4 & +9.5\% \\
Micro-F1 & 48.5\% & 55.6\% & +7.1 & +14.6\% \\
Macro-F1 & 32.1\% & 36.9\% & +4.8 & +15.0\% \\
Weighted-F1 & 46.8\% & 55.1\% & +8.3 & +17.7\% \\
Samples-F1 & 41.2\% & 47.1\% & +5.9 & +14.3\% \\
Mean AUROC & 61.3\% & 68.9\% & +7.6 & +12.4\% \\
\midrule
\textbf{Mean Gain} & & & \textbf{+6.7 pp} & \textbf{+13.9\%} \\
\bottomrule
\end{tabular}
\end{table}

The relative improvements reveal interesting patterns in how LoRA affects different aspects of classification performance. Weighted-F1 shows the largest relative gain at +17.7\%, followed closely by macro-F1 at +15.0\% and micro-F1 at +14.6\%. These substantial F1 improvements, particularly for macro-F1, suggest that LoRA fine-tuning disproportionately benefits underrepresented pathologies where the base model struggles. This is clinically significant because rare conditions often receive less attention during pretraining due to class imbalance, yet they frequently represent the most diagnostically challenging cases. The relatively modest accuracy gain of +9.5\%, compared to the F1 improvements, indicates that the model becomes better at handling difficult positive cases rather than simply improving on easy negatives. Mean AUROC improvement of +7.6 pp confirms that the enhanced representations maintain discriminative power across varying decision thresholds, not just at a single operating point. Across all metrics, the mean relative improvement of +13.9\% demonstrates that even with only 0.38\% of parameters being trained, LoRA captures task-relevant features that substantially improve the model's ability to align CT volumes with textual pathology descriptions.

\section{Conclusions}
In this work, we introduced MedCT-VLM, a parameter-efficient vision–language framework designed to adapt large-scale CT foundation models for downstream clinical applications. We demonstrated that LoRA adapters enable efficient fine-tuning of CT-CLIP for multi-label thoracic pathology classification. By inserting adapters into both vision and text encoders while freezing the base model, we improved zero-shot AUROC from 61.3\% to 68.9\% across 18 pathologies, with consistent gains in accuracy and F1. The approach reduced checkpoint size by 74$\times$ while training only 0.38\% of parameters. Future work will evaluate performance on larger datasets and assess clinical utility in real-world settings.

\bibliographystyle{elsarticle-num}
\bibliography{refs}

\end{document}